\documentclass{article}

\usepackage[nonatbib,preprint]{iaseai26}

\usepackage[utf8]{inputenc} % allow utf-8 input
\usepackage[T1]{fontenc}    % use 8-bit T1 fonts
\usepackage{hyperref}       % hyperlinks
\usepackage{url}            % simple URL typesetting
\usepackage{booktabs}       % professional-quality tables
\usepackage{amsfonts}       % blackboard math symbols
\usepackage{nicefrac}       % compact symbols for 1/2, etc.
\usepackage{microtype}      % microtypography
\usepackage{xcolor}         % colors
\usepackage{amsmath, amssymb, dsfont}
\usepackage{booktabs, colortbl, graphicx, ragged2e}
\usepackage{siunitx, array, multirow} % for number alignment
\definecolor{lightgreen}{rgb}{0.88,1,0.88}
\usepackage{subcaption}
\usepackage{enumitem}
\setlist[itemize]{leftmargin=1.5em}

\usepackage[most]{tcolorbox}

\usepackage[backend=biber,style=numeric,sorting=none,defernumbers=true]{biblatex}

\title{KnowRL: Teaching Language Models to Know What They Know}

\author{%
  Sahil Kale \\
  KnowledgeVerse AI\\
  Atlanta, Georgia \\
  USA \\
  \And
  Devendra Singh Dhami \\
  Department of Mathematics and Computer Science \\
  Uncertainty in Artificial Intelligence Group \\
  TU Eindhoven, Eindhoven, The Netherlands \\
}

\begin{document}

\maketitle

\begin{abstract}
  Truly reliable AI requires more than simply scaling up knowledge; it demands the ability to know what it knows and when it does not. Yet recent research shows that even the best LLMs misjudge their own competence in more than one in five cases, making any response born of such internal uncertainty impossible to fully trust. Inspired by self-improvement reinforcement learning techniques that require minimal data, we present a simple but powerful framework \textit{KnowRL} that strengthens a model’s internal understanding of its own feasibility boundaries, enabling safer and more responsible behaviour. Our framework combines two components: (i) introspection, where the model generates and classifies tasks it judges feasible or infeasible, and (ii) consensus-based rewarding, where stability of self-knowledge assessment is reinforced through internal agreement. By using internally-generated data, this design strengthens consistency in self-knowledge and entirely avoids costly external supervision. In experiments on LLaMA-3.1-8B and Qwen-2.5-7B, \textit{KnowRL} steadily improved self-knowledge, validated by both intrinsic self-consistency and extrinsic benchmarking. With nothing more than a small seed set and no external supervision, our method drove gains as high as 28\% in accuracy and 12\% in F1, outperforming baselines in just a few iterations. Our framework essentially unlocks the untapped capacity of LLMs to self-improve their knowledge awareness, opening the door to reliable, more accountable AI and safer deployment in critical applications. Owing to its simplicity and independence from external effort, we encourage applying this reliability-enhancing process to all future models, and we release all code and data publicly to support broad adoption \footnote{\url{https://anonymous.4open.science/r/KnowRL-5BF0}}.
\end{abstract}

\section{Introduction}
True intelligence is measured not just by the accumulation of knowledge, but by the ability to recognise the limits of our own understanding \cite{liang-etal-2024-learning,ren2024investigatingfactualknowledgeboundary}. In terms of large language models (LLMs), this refers to the fundamental property of self-knowledge, defined as the ability for models to clearly delineate between feasible and infeasible tasks based on knowing their own capability and knowledge boundaries \cite{wang-etal-2023-self-knowledge}. Recent research demonstrates that even leading models misjudge their competence in more than one out of five instances \cite{kale-vrn-2025-line}, leading to severe trust and safety issues since responses drawn from such internal uncertainty can never be considered completely reliable \cite{ni2024llmsneedretrievalaugmentation}. Methods that enable LLMs to reliably and consistently recognise the boundaries of their own knowledge are urgent not only for ensuring true reliability and trustworthiness, but also to enable widespread AI adoption, safely.

The problem of self-knowledge, or the lack thereof in LLMs, has been highlighted as a stand-alone issue \cite{ren2024investigatingfactualknowledgeboundary,yin-etal-2023-large,song2025language}, an issue related to other known problems like memorisation \cite{kale2025miragemasterymemorizationtricks,satvaty2025undesirablememorizationlargelanguage} and adversarial helpfulness \cite{ajwani2024llmgeneratedblackboxexplanationsadversarially}, and also as an underlying cause for AI safety issues \cite{griot2025llm_metacognition,zhang2025safetyrefusalreasoningenhancedfinetuning}. Since external databases or scaffolding techniques are not suitable to resolve this ingrained gap \cite{chang2024external}, modifying training patterns or applying post-hoc reinforcement and fine-tuning remain the most effective and viable approaches to address the problem. Calibrating a model’s confidence can signal when an answer might be wrong, but it does not guarantee that the model is consistent about what it truly knows versus what it doesn’t. We therefore, like previous research, treat self-knowledge as a separate problem from uncertainty estimation. 

In this paper, we leverage Reinforcement Learning (RL), a powerful alignment strategy \cite{liu2025survey}, to systematically reinforce self-knowledge within large language models. Since models show large wavering in their own perception of knowledge boundaries \cite{kale-vrn-2025-line}, we introduce a self-consistency based approach that explicitly strengthens the model’s understanding of feasibility and infeasibility limits for itself. We posit that the capacity for such introspection already exists within LLMs, and our approach serves to reinforce and guide this latent ability, empowering the models to better understand and articulate what they do and do not know. By using a variation on Self-Play Reinforcement Learning \cite{fang2025serlselfplayreinforcementlearning}, our method enables LLMs to bootstrap a more accurate and consistent view of their own boundaries, even with minimal initial data. 

\begin{figure}[t]
\begin{center}
  \includegraphics[width=0.4\columnwidth]{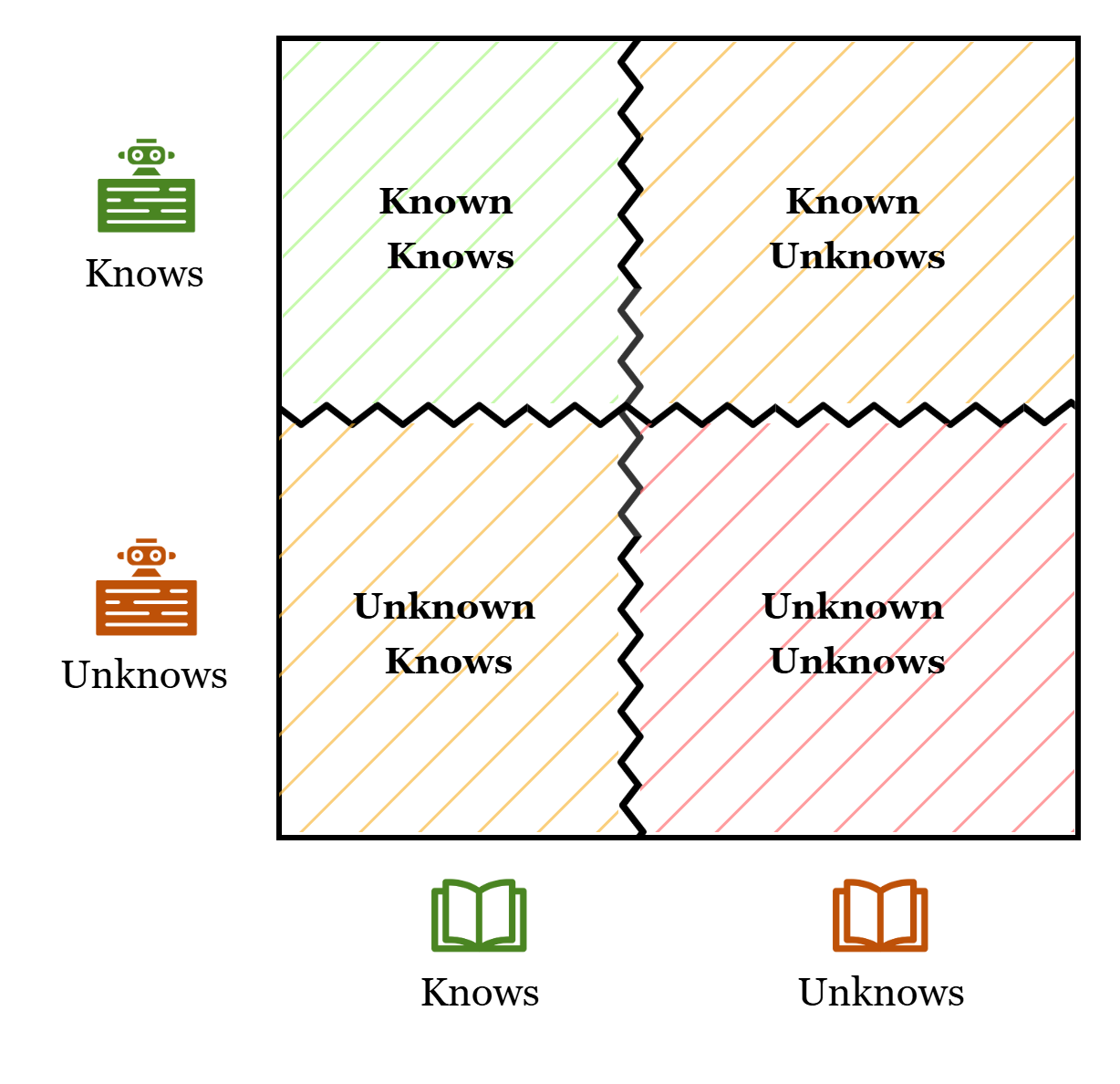}
  \caption{The main idea of self-knowledge represented by the know-unknow quadrant, adapted from \cite{yin-etal-2023-large}. The horizontal axis reflects the model’s awareness of the information space, while the vertical axis captures the model’s capacity to accurately comprehend and apply what it knows.}
  \label{fig:intro}
\end{center}
\end{figure}

Since current large language models show a significant lack of self-knowledge, their responses can never be considered inherently reliable and perfectly grounded \cite{ni2024llmsneedretrievalaugmentation}. To represent visually, Figure \ref{fig:intro} depicts self-knowledge as the know-unknow quadrant, with jagged boundaries reflecting uncertainty within this distinction in LLMs. Our goal is to smooth and formalise these boundaries so LLMs can recognize their own knowledge limits and adapt their responses accordingly, enabling a new class of responsible, text-based AI systems. We believe this can also address known problems like over-refusal \cite{sullutrone-etal-2025-cover}, sycophancy \cite{rrv-etal-2024-chaos} and adversarial helpfulness \cite{ajwani2024llmgeneratedblackboxexplanationsadversarially}. Beyond technical improvements, this work has important societal implications for safer AI deployment. We encourage researchers to adopt our simple post-training reinforcement framework across critical areas like healthcare, law, and finance, where trustworthy and accountable AI is essential for effective decision-making. 

We make our method advantageous and simple to use by inherently addressing two key challenges. First, collecting high-quality human-annotated data in most domains is expensive and difficult \cite{kandpal2025positionexpensivellmtraining}, but since our goal is to enhance internal knowledge awareness, using LLM-generated data proves both effective and better aligned with our objective. Second, using single model-generated responses directly as ground-truth labels can produce unreliable reward signals in reinforcement learning \cite{liu2024elephantroomunveilingimpact}. We address this by implementing a self-rewarding mechanism based on  consensus \cite{zuo2025ttrltesttimereinforcementlearning} and consistency of feasibility, which provides stable, trustworthy, yet internally-generated signals. Our main contributions are summarised as follows:

\begin{enumerate}
    \item We explore how to leverage RL to guide introspection in LLMs for enhanced awareness of self-knowledge boundaries to enable safer AI, even with limited initial data and no external supervision.
    \item We introduce the KnowRL framework, built on two key components: introspection, where the LLM generates questions it judges as feasible or infeasible, and consensus-based rewarding, which derives stable, trustworthy reward signals from internal agreement to reinforce the model’s self-knowledge.
    \item We show that KnowRL can strongly boost LLM self-knowledge, achieving up to 28\% accuracy and 12\% F1 gains in just a few iterations, demonstrating scalable self-improvement with broader implications for safer, more reliable AI.
\end{enumerate}

\section{Preliminaries}
\subsection{Reinforcement Learning in LLMs}
Reinforcement learning (RL) refers to a framework where an agent interacts with an environment by taking actions, receiving rewards, and updating its policy to maximise long-term expected returns \cite{hu-etal-2018-playing}. In the context of large language models, RL has been widely applied to adjust model outputs to human preferences, for post-hoc alignment through human feedback like RLHF \cite{NIPS2017_d5e2c0ad}. More recently, it has also been explored for improving factuality \cite{chen2025learningreasonfactuality} and AI safety \cite{mu2024rulebasedrewardslanguage}, demonstrating versatility in refining LLM behaviour beyond pure pre-training.

Language modelling can be directly interpreted as an RL problem, since generating a single response is equivalent to taking an action, and the quality of that response can be analysed to get a reward signal and improve the model. For an input $x$, a policy $\pi_\theta(y \mid x)$ generates a response $y$, and the model receives a reward based on $y^*$, the reference answer. 
\begin{equation}
    r = R(x, y, y^*),
    \label{eq:reward}
\end{equation}

The training objective is to maximise the expected reward:
\begin{equation}
    J(\theta) = \mathbb{E}_{y \sim \pi_\theta(\cdot \mid x)}\big[ R(x, y, y^*) \big].
    \label{eq:objective}
\end{equation}

In this view, generating a response is the action, the reward reflects response quality, and optimisation adjusts $\pi_\theta$ to favour better responses.

Traditionally, rewards rely on reference-based similarity, such as parsers or verifiers \cite{zhang2025generativeverifiersrewardmodeling}. More recent work removes the dependency on $y^*$ by introducing unsupervised rewards, including majority voting \cite{wei2025unsupervisedposttrainingmultimodalllm} or divergence-based measures \cite{wang2023reverseklgeneralizingdirect}. 

In our context, we focus on reinforcing \textit{awareness of self-knowledge boundaries} more than quality or alignment of responses. By treating precise and consistent self-knowledge estimation as the rewarded behaviour, we can apply RL not just to improve correctness, but to strengthen a model’s ability to recognise and respect the boundary between what it knows and what it does not, based on its own internal knowledge.

\begin{figure*}[t]
\begin{center}
  \includegraphics[width=0.95\columnwidth]{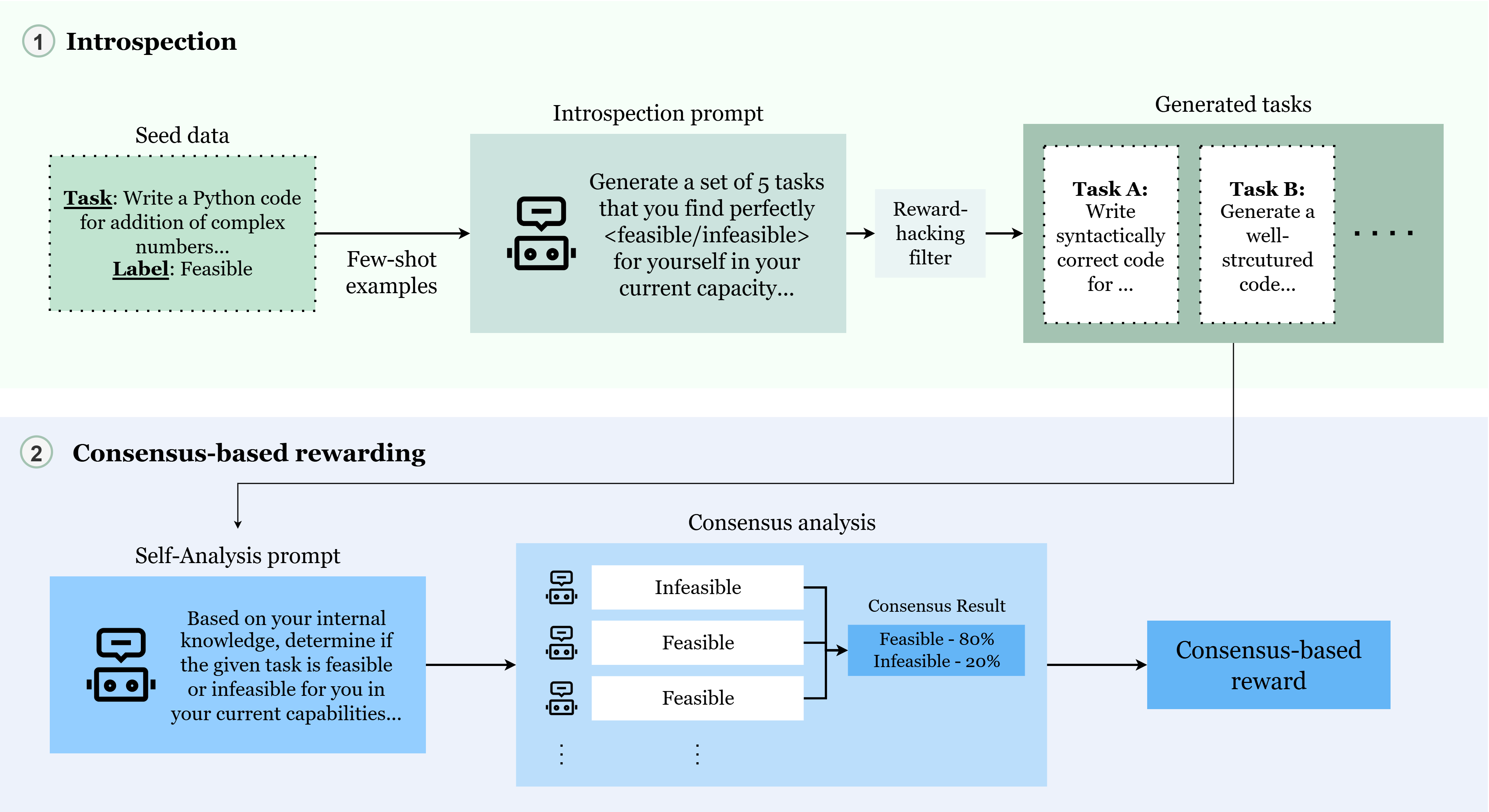}
  \caption{The KnowRL framework, with two key components in an iterative, RL-driven training loop}
  \label{fig:methodology}
\end{center}
\end{figure*}

\subsection{Self-Play Reinforcement Learning}

Self-play reinforcement learning \cite{fang2025serlselfplayreinforcementlearning} extends the standard RL setup by enabling a model to generate both the inputs and the corresponding learning rewards. Rather than depending on external supervision, the model interacts with its own outputs to create a feedback loop, which is then used to update its policy. This paradigm has been widely applied in domains such as games \cite{silver2017masteringchessshogiselfplay} and robotics \cite{openai2021asymmetricselfplayautomaticgoal}, and more recently explored in language modelling as well \cite{ji2024selfplayadversarialcriticprovable,selfplayllm}.

In general, for an input $x$ generated through self-play, a policy $\pi_\theta(y \mid x)$ produces a response $y$. A reward signal $r = R(x, y)$ is then derived from internal consistency or agreement, rather than a fixed reference $y^*$. The objective remains to maximise expected reward:
\begin{equation}
    J(\theta) = \mathbb{E}_{y \sim \pi_\theta(\cdot \mid x)}[R(x, y)].
    \label{eq:selfplayobjective}
\end{equation}

In our setting, self-play is not used to optimise for raw task accuracy, but to reinforce self-knowledge. Our \textsc{KnowRL} framework operates in two stages. In the introspection step, the LLM generates feasible or infeasible questions and tasks for itself, to probe the limits of its own knowledge. Second, a consensus-based rewarding step, where multiple judgments of its self-generated tasks $\{y_i\}_{i=1}^k$ are aggregated, and the reward is defined as the quantified consistency in agreement, producing a stable and trustworthy signal without external labels.

\begin{equation}
    r = Consistency\big(\{y_i\}_{i=1}^k\big),
    \label{eq:majorityreward}
\end{equation}

Our method thus embeds the loop entirely within a single LLM. By combining introspection with consensus-driven rewards, we obtain a lightweight and scalable strategy for reinforcing reliable boundary awareness, crucial for trust and safety in downstream deployment.

\begin{table}[ht]
\centering
\caption{Results of intrinsic evaluation of self-knowledge boundary consistency for KnowRL across iterations. $\Delta$ indicates change from the previous step.}
\vspace{0.5em}
\renewcommand{\arraystretch}{1.2}
\small
\begin{tabular}{>{\centering\arraybackslash}m{4cm} >{\centering\arraybackslash}m{2.5cm} 
                >{\centering\arraybackslash}m{2.5cm} >{\centering\arraybackslash}m{2.5cm}}
\toprule
Model & Iteration & Accuracy (\%) & $\Delta$ (\%) \\
\midrule
\multirow[c]{7}{*}{Llama 3.1 8B Instruct} 
  & \cellcolor{lightgreen!0} Base Model & \cellcolor{lightgreen!0} 33.56 & \cellcolor{lightgreen!0} - \\
  & \cellcolor{lightgreen!20} Iter 5  & \cellcolor{lightgreen!20} 36.78 & \cellcolor{lightgreen!20} 3.22 $\uparrow$ \\
  & \cellcolor{lightgreen!30} Iter 10 & \cellcolor{lightgreen!30} 39.32 & \cellcolor{lightgreen!30} 2.54 $\uparrow$ \\
  & \cellcolor{lightgreen!40} Iter 15 & \cellcolor{lightgreen!40} 41.11 & \cellcolor{lightgreen!40} 1.79 $\uparrow$ \\
  & \cellcolor{lightgreen!50} Iter 20 & \cellcolor{lightgreen!50} 42.13 & \cellcolor{lightgreen!50} 1.02 $\uparrow$ \\
  & \cellcolor{lightgreen!60} Iter 25 & \cellcolor{lightgreen!60} 43.12 & \cellcolor{lightgreen!60} 0.99 $\uparrow$ \\
  & \cellcolor{lightgreen!50} Iter 30 & \cellcolor{lightgreen!50} 42.99 & \cellcolor{lightgreen!50} 0.13 $\downarrow$ \\
\midrule
\multirow[c]{7}{*}{Qwen 2.5 7B Instruct} 
  & \cellcolor{lightgreen!0} Base Model & \cellcolor{lightgreen!0} 39.22 & \cellcolor{lightgreen!0} - \\
  & \cellcolor{lightgreen!20} Iter 5  & \cellcolor{lightgreen!20} 43.19 & \cellcolor{lightgreen!20} 3.97 $\uparrow$ \\
  & \cellcolor{lightgreen!30} Iter 10 & \cellcolor{lightgreen!30} 45.78 & \cellcolor{lightgreen!30} 2.59 $\uparrow$ \\
  & \cellcolor{lightgreen!40} Iter 15 & \cellcolor{lightgreen!40} 46.71 & \cellcolor{lightgreen!40} 0.93 $\uparrow$ \\
  & \cellcolor{lightgreen!50} Iter 20 & \cellcolor{lightgreen!50} 46.77 & \cellcolor{lightgreen!50} 0.06 $\uparrow$ \\
  & \cellcolor{lightgreen!60} Iter 25 & \cellcolor{lightgreen!60} 48.01 & \cellcolor{lightgreen!60} 1.24 $\uparrow$ \\
  & \cellcolor{lightgreen!50} Iter 30 & \cellcolor{lightgreen!50} 48.29 & \cellcolor{lightgreen!50} 0.28 $\uparrow$ \\
\bottomrule
\end{tabular}
\label{tab:intr}
\end{table}

\section{Methodology}
To address the persistent wavering in large language models’ perception of their own knowledge boundaries \cite{wang-etal-2023-self-knowledge,kale-vrn-2025-line}, we propose a self-consistency driven framework that reinforces an LLM’s internal understanding of feasibility and infeasibility limits. Building on the frameworks of self-play \cite{fang2025serlselfplayreinforcementlearning} and self-improvement \cite{chen2025selfquestioninglanguagemodels}, our approach enables models to bootstrap a more accurate and stable view of their own boundaries, even with minimal initial data and external supervision. We describe the \textsc{KnowRL} framework described in Figure \ref{fig:methodology} in detail ahead, highlighting how the two components: \textit{introspection} and \textit{consensus-based rewarding} interact to drive iterative improvement in self-knowledge.

\subsection{Introspection}
Our primary goal is to strengthen a language model’s ability to recognise the limits of its own knowledge and capabilities. To achieve this, we design a method that relies on and utilises data generated by the model itself, minimising external supervision. In the introspection step, the model is prompted to propose tasks it confidently believes are either feasible or infeasible, guided by a small set of few-shot verified examples (see Figure \ref{fig:methodology}). As training steps progress, we utilise examples from the initial dataset as well as samples with high consensus generated in previous training steps (examples in Section \ref{sec:appendix-eg} in the Appendix). We believe that incorporating such samples can allow the model to refine, evolve and stabilise its understanding of feasibility boundaries over successive self-improvement iterations.

\subsection{Consensus-based Rewarding}
In the consensus-based rewarding stage, the model’s own judgments are used to quantify and reinforce the consistency of its self-knowledge. Our main motivations and goal in this stage is to reward model introspection that produces high consensus and consistency and leads to a strong and rooted understanding of its own feasibility boundaries. Formally, given a candidate task $x$ produced during introspection, we draw $k$ independent self-analysis outputs $\{y_i\}_{i=1}^{k}$, where each $y_i \in \{\text{Feasible}, \text{Infeasible}\}$, using the analysis prompt given in Figure \ref{fig:rl-p2}. Within the prompt, we use simple strategies like QAP \cite{yugeswardeenoo-etal-2024-question} and chain-of-thought \cite{wei2023chainofthoughtpromptingelicitsreasoning} to encourage better understanding before coming to a conclusion of task feasibility. 

The reward is the proportion of outputs agreeing with the majority label, which directly measures the internal consistency of feasibility assessments.  
\[
r(x) = \frac{1}{k} \sum_{i=1}^{k} \mathbf{1}\Big[y_i = \text{Maj}\{y_1, \dots, y_k\}\Big],
\]

This consensus score serves as the reinforcement signal in our policy update, encouraging the model to generate tasks and judgments that reflect a stable and reliable perception of its own capability boundaries.  

\begin{table}[ht]
\centering
\caption{Results of extrinsic evaluation of self-knowledge boundary consistency using the SelfAware dataset for KnowRL across iterations. $\Delta$ indicates change from the previous step.}
\vspace{0.5em}
\renewcommand{\arraystretch}{1.2}
\small
\begin{tabular}{>{\centering\arraybackslash}m{4cm} >{\centering\arraybackslash}m{2.5cm} 
                >{\centering\arraybackslash}m{2.5cm} >{\centering\arraybackslash}m{2.5cm}}
\toprule
Model & Iteration & F1 (\%) & $\Delta$ (\%) \\
\midrule
\multirow[c]{7}{*}{Llama 3.1 8B Instruct} 
  & \cellcolor{lightgreen!0} Base Model & \cellcolor{lightgreen!0} 56.12 & \cellcolor{lightgreen!0} - \\
  & \cellcolor{lightgreen!20} Iter 5  & \cellcolor{lightgreen!20} 58.01 & \cellcolor{lightgreen!20} 1.89 $\uparrow$ \\
  & \cellcolor{lightgreen!30} Iter 10 & \cellcolor{lightgreen!30} 58.65 & \cellcolor{lightgreen!30} 0.64 $\uparrow$ \\
  & \cellcolor{lightgreen!40} Iter 15 & \cellcolor{lightgreen!40} 61.76 & \cellcolor{lightgreen!40} 3.11 $\uparrow$ \\
  & \cellcolor{lightgreen!50} Iter 20 & \cellcolor{lightgreen!50} 62.34 & \cellcolor{lightgreen!50} 0.58 $\uparrow$ \\
  & \cellcolor{lightgreen!60} Iter 25 & \cellcolor{lightgreen!60} 63.11 & \cellcolor{lightgreen!60} 0.77 $\uparrow$ \\
  & \cellcolor{lightgreen!50} Iter 30 & \cellcolor{lightgreen!50} 63.10 & \cellcolor{lightgreen!50} 0.01 $\downarrow$ \\
\midrule
\multirow[c]{7}{*}{Qwen 2.5 7B Instruct} 
  & \cellcolor{lightgreen!0} Base Model & \cellcolor{lightgreen!0} 62.17 & \cellcolor{lightgreen!0} - \\
  & \cellcolor{lightgreen!20} Iter 5  & \cellcolor{lightgreen!20} 64.88 & \cellcolor{lightgreen!20} 2.71 $\uparrow$ \\
  & \cellcolor{lightgreen!30} Iter 10 & \cellcolor{lightgreen!30} 65.94 & \cellcolor{lightgreen!30} 1.06 $\uparrow$ \\
  & \cellcolor{lightgreen!40} Iter 15 & \cellcolor{lightgreen!40} 67.22 & \cellcolor{lightgreen!40} 1.28 $\uparrow$ \\
  & \cellcolor{lightgreen!50} Iter 20 & \cellcolor{lightgreen!50} 67.89 & \cellcolor{lightgreen!50} 0.67 $\uparrow$ \\
  & \cellcolor{lightgreen!60} Iter 25 & \cellcolor{lightgreen!60} 68.23 & \cellcolor{lightgreen!60} 0.34 $\uparrow$ \\
  & \cellcolor{lightgreen!50} Iter 30 & \cellcolor{lightgreen!50} 68.29 & \cellcolor{lightgreen!50} 0.06 $\uparrow$ \\
\bottomrule
\end{tabular}
\label{tab:extr}
\end{table}

\subsection{Overall Setup}
The KnowRL framework integrates introspection and consensus-based rewarding in an iterative, RL-powered training loop to progressively improve the model’s understanding of its own capabilities. During each introspection phase, the model is prompted to generate candidate tasks it believes are either feasible or infeasible, with each introspection run repeated $10$--$15$ times to produce a diverse set of approximately $50$--$60$ candidate tasks. For each candidate task $x$, we draw $k=8$ independent self-analysis outputs $\{y_i\}_{i=1}^{k}$, where $y_i \in \{\text{Feasible}, \text{Infeasible}\}$, and compute a consensus-based reward to quantify the internal consistency of the model’s feasibility judgments. The set of tasks and corresponding rewards, $\{(x, r(x))\}$, are then used to update the model parameters via reinforcement learning. We refer to one iteration or cycle as running the introspection and consensus-rewarding process once for feasible tasks and once for infeasible tasks, with each run used to update the model through reinforcement learning. By repeating this cycle iteratively across training steps, the LLM progressively strengthens using our RL-powered introspective loop, improving its ability to recognise and articulate the limits of what it can or cannot accomplish without any external labels.

\subsection{Reward Hacking Filter}
During spot checks across intermediate iterations, we observed  that models tend to produce overly simple or overly complex tasks once they realise that consensus agreement is more when certain keywords or phrasing is used, essentially hacking the consensus-based reward. To prevent such hacking, where the model learns to maximise consensus by generating overly simple or unnecessarily complex tasks, we adopt a difficulty clipping strategy inspired by prior self-play methods \cite{fang2025serlselfplayreinforcementlearning}. During training, tasks produced in the introspection step are filtered out based on predetermined criteria.

\begin{itemize}
    \item \textbf{Semantic redundancy:} If the ROUGE-L score for any new task exceeds a predefined threshold with existing instructions, we filter out such examples to prevent generating semantically similar instructions that reduce diversity.
    \item \textbf{Keyword filtering:} The presence of specific keywords such as \textit{generating images}, \textit{training models}, or \textit{image, video}, which can directly lead to infeasibility consensus agreement owing to model capability limitations are filtered. We also add this filter since awareness of text-only capabilities are pretty high even without our RL-powered tuning.
    \item \textbf{Perplexity filtering:} We utilise the negative log-likelihood under the base model and discard candidates with perplexity above a fixed threshold, ensuring syntactically fluent and semantically well-formed instructions.
    
\end{itemize}
These filters maintain a balanced level of task difficulty and prevent the model from exploiting consensus signals through trivial patterns or unanswerable prompts.

\begin{figure}[ht]
    \centering
    % Subfigure 1
    \begin{subfigure}[b]{0.48\textwidth}  % width can be adjusted
        \centering
        \includegraphics[width=\linewidth]{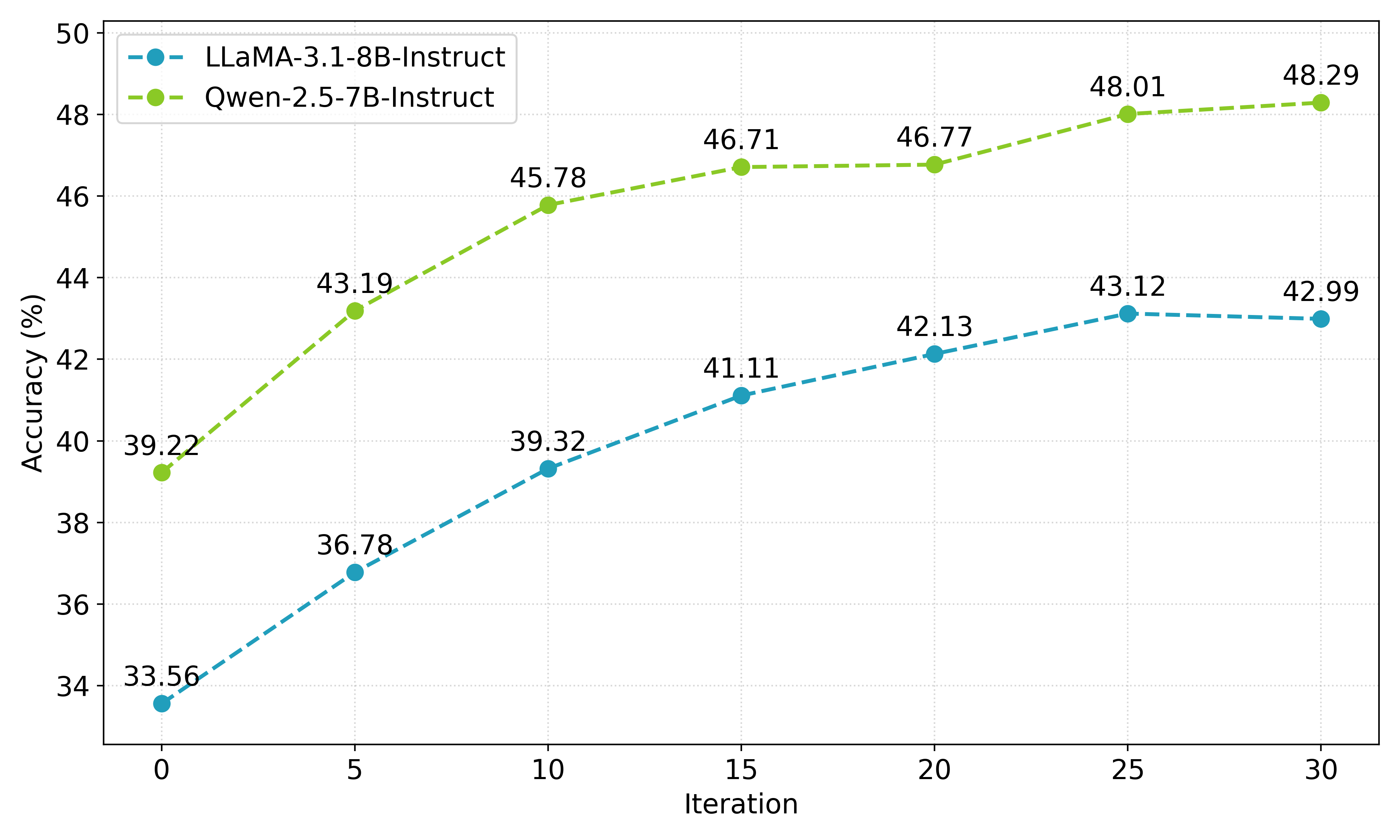}  % replace with your file
        \caption{Accuracy trend of intrinsic evaluation \cite{kale-vrn-2025-line} for both models using KnowRL}
        \label{fig:improve-intr}
    \end{subfigure}
    \hfill
    % Subfigure 2
    \begin{subfigure}[b]{0.48\textwidth}
        \centering
        \includegraphics[width=\linewidth]{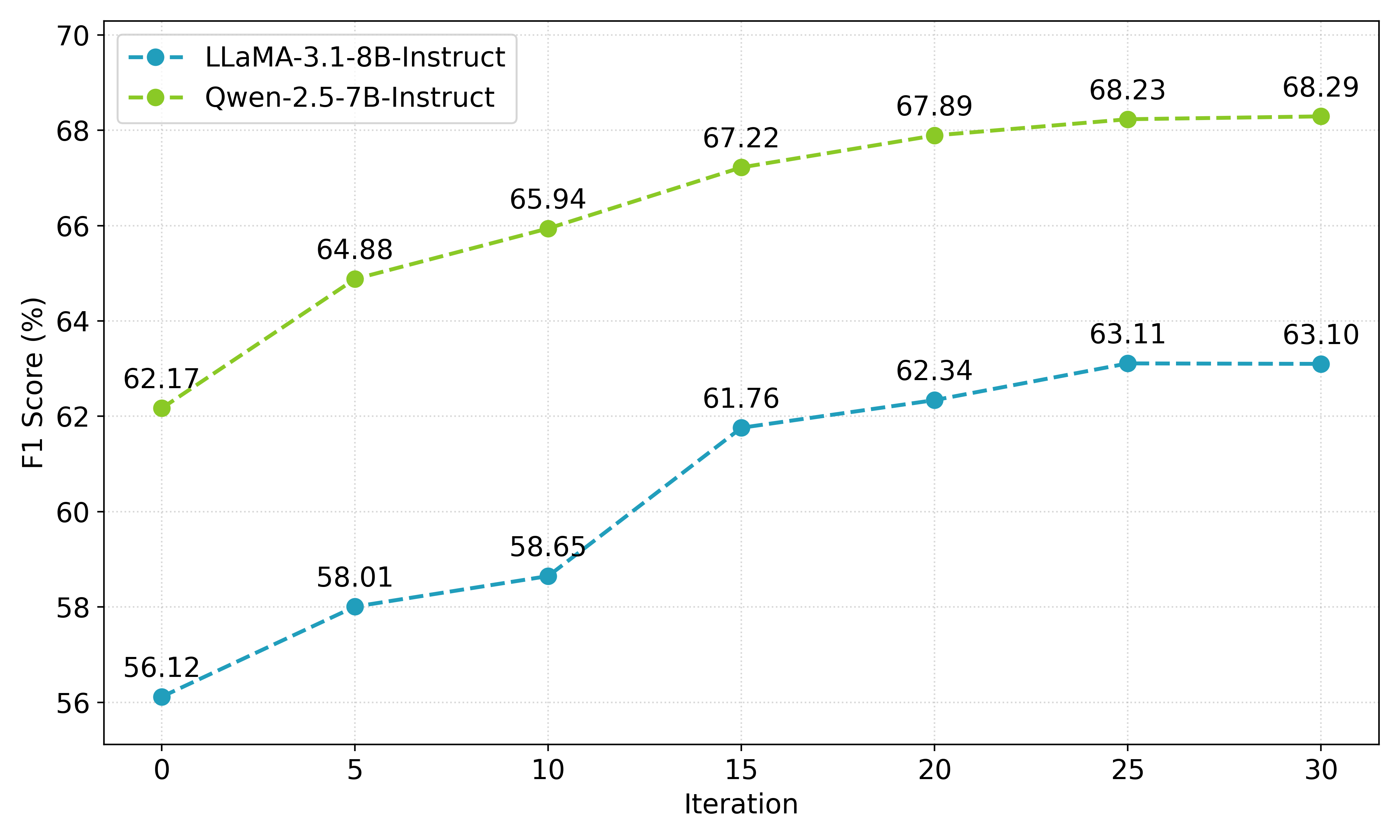}  % replace with your file
        \caption{F1 score trend of extrinsic evaluation using the SelfAware dataset \cite{yin-etal-2023-large} for both models using KnowRL}
        \label{fig:improve-extr}
    \end{subfigure}
    \caption{Evaluation of self-knowledge boundary consistency}
    \label{fig:improve}
\end{figure}

\section{Experimentation}
\subsection{Model Setup}
We conduct our experimentation with two small yet powerful open-source models: Llama 3.1 8B Instruct \cite{meta_llama_3_1_8b_instruct} and Qwen-2.5-7B-Instruct \cite{qwen2025qwen25technicalreport}. Model parameters and details are found in the Appendix. For both models, we use a seed dataset as few-shot examples for the introspection step containing verified answerable and unanswerable questions generated by the models themselves. More details and prompts for the seed dataset are provided in Section \ref{sec:appendix-seed} in the Appendix. For training, we use the OpenRLHF \cite{hu2025openrlhfeasytousescalablehighperformance} framework together with its Reinforce++ algorithm \cite{hu2025reinforceefficientrlhfalgorithm} to perform reinforcement learning. Detailed training hyperparameters during the RL process are provided in Section \ref{sec:appendix-a2} in the Appendix.

\subsection{Dataset and Evaluation}
To comprehensively assess our methodology, we employ two complementary evaluation settings: intrinsic analysis and dataset-based evaluation. We run RL-based introspection with consensus rewarding for up to 30 iterations, check-pointing the model and evaluating its performance every 5 iterations. Given the lack of established methods targeting intrinsic self-knowledge improvement for comparison, we treat the base model’s performance as the baseline for evaluation.

\noindent \textbf{Intrinsic Evaluation:} Since self-knowledge is an inherent property of language models, we agree with \cite{kale-vrn-2025-line} that evaluation using self-generated data is most suitable and provides a grounded view. Specifically, we adopt their generation–validation consistency method to assess the accuracy and consistency of the model’s perception of their own self-knowledge boundaries. For each evaluation, we repeat the generation and validation process 250 times for both feasible and infeasible tasks using the prompts for the vanilla setup given in the original paper, and report the average accuracy as the primary evaluation metric. Results for both models are shown in Table \ref{tab:intr}.

\noindent \textbf{Extrinsic Evaluation:}
To further validate self-knowledge gains using our technique, we evaluate the model on a standard external benchmark: the SelfAware dataset \cite{yin-etal-2023-large}. This dataset contains a collection of answerable and unanswerable questions along with explanations for unanswerability. We randomly sample 500 questions of each type and perform inference using the in-context learning-based prompt provided in the original paper. Performance is measured using the F1 score, which provides a balanced view of precision and recall. Results obtained from both models for extrinsic evaluation are shown in Table \ref{tab:extr}.

\section{Results and Discussion}
\textbf{Significant self-knowledge improvement across both models validates RL framework effectiveness} : Both Qwen 7B and LLaMA 8B achieved notable gains in self-knowledge over their baselines. Intrinsic evaluation showed over 28\% accuracy improvement for LLaMA and around 23\% for Qwen across 30 training iterations, while extrinsic evaluation on the SelfAware benchmark recorded around 10 and 12\% F1 gains over the base models, respectively. We emphasise that these improvements were achieved using only a \textit{small seed dataset and no external annotations during training}, underscoring the efficiency and scalability of our approach, and showing that the framework can scale to all domains lacking labelled data. This demonstrates that our RL-driven introspection can consistently strengthen self-knowledge boundaries across model sizes, leading to more responsible and trustworthy outputs even without any external labels. Our method enables safer, deployment-ready models for high-stakes domains like healthcare, law, education, and science, where unchecked over- or under-confidence carries serious risk.

\noindent\textbf{Steady, monotonic gains in self-knowledge across training iterations highlight self-improvement capacity:} Both models showed clear, monotonic improvement at nearly every checkpoint (as seen in Figure \ref{fig:improve}), reflecting a stable internal growth in understanding their own feasibility boundaries. These results indicate that language models inherently possess the capacity to refine their self-knowledge, and our RL framework effectively reinforces this awareness and understanding of their own feasibility boundaries for themselves. Since maximal improvement can be seen with minimal data and a few initial training cycles, self-knowledge improvements can be made cheap, predictable, and efficient. However, we also acknowledge that progress began to level off around iterations 25–30, beyond which we could not test due to computational constraints, suggesting a natural limit to internal self-improvement or the need for stronger signals to push beyond the observed plateau.

\noindent\textbf{Current LLMs remain unsafe for critical deployment without interventions, but self-improvement offers real hope:} Our results reveal that base models start with alarmingly low self-knowledge, a weakness that directly undermines AI safety in almost all domains. Such fragile awareness means today’s LLMs are not yet ready for deployment to reliably and transparently help humans. Yet, our method shows that even simple reinforcement-based self-improvement can drive steady, reliable gains in a core safety property. This provides a concrete path forward, since similar frameworks based on self-improvement and reinforcing internal capabilities and awareness could target other persistent issues such as memorisation, sycophancy, and data–instruction boundary failure to lead to safer AI models moving forward.

\section{Related Work}
\textbf{Self-Knowledge in LLMs:} The problem of wavering and inconsistent self-knowledge has been highlighted in a large body of work on AI safety in recent times. Studies have taken several approaches to formalise and improve this inherent non-trivial problem: dataset-based evaluation of binary classification results of answerability \cite{zhang2024largelanguagemodelsgood,wen2024perceptionknowledgeboundarylarge}, intrinsic evaluation based on internal consistency \cite{kale-vrn-2025-line,li2023benchmarkingimprovinggeneratorvalidatorconsistency}, and self-cognition \cite{chen2024selfcognitionlargelanguagemodels}. Methods such as Self-Reflect \cite{kirchhof2025selfreflectiveuncertaintiesllmsknow} which summarise the model’s own internal answer distribution, or uncertainty-aware instruction tuning that trains models to express uncertainty more effectively, have been possible techniques to enhance the \textit{unknowing} part of self-knowledge. In contrast, our work aims not merely to measure or expose inconsistency, but to smooth and formalise these feasibility boundaries within LLMs, so that the model itself can track and adapt to its knowledge limits, enabling safer and more reliable behaviour.

\noindent\textbf{Self-Improvement in LLMs:} The capacity for LLMs to improve themselves with minimal external support has been demonstrated in a number of recent works. For instance, Self-Refine \cite{madaan2023selfrefineiterativerefinementselffeedback} lets an LLM generate an initial answer, then critique and iteratively refine it using its own feedback, while \cite{huang2022largelanguagemodelsselfimprove} make the model use chain-of-thought prompting and self-consistency to produce rationale-augmented answers for unlabelled inputs. Also, synthetic data-based methods methods like Self-Taught Evaluator \cite{wang2024selftaughtevaluators} and K2 \cite{liu2025llm360k2building65b} use self-generated training set of reasoning tasks and answers for improving performance. On the RL front, there are approaches that use post-hoc reinforcement or reward signals to sharpen model behaviour. RLRF \cite{lee2024reinforcementlearningreflectivefeedback} has the model reflect on its responses using fine-grained criteria, while R-Zero \cite{huang2025rzeroselfevolvingreasoningllm}, SeRL \cite{fang2025serlselfplayreinforcementlearning} and Self-Questioning \cite{chen2025selfquestioninglanguagemodels} use two models or agents that co-evolve with RL to ensure improvement without human labels. Taking inspiration from such successful techniques, we show how LLMs can progressively tighten their self-knowledge boundaries as well using self-improvement.

\section{Conclusion}
In this work, we introduce a self-consistency driven RL framework to improve large language models’ self-knowledge, addressing their persistent difficulty in accurately and consistently being aware of the limits of their own capabilities, to improve response trustworthiness and safety. By rewarding the generation of introspective problems and reinforcing awareness via internal agreement, our approach enables models to bootstrap a stable understanding of feasible and infeasible tasks using only self-generated data and minimal supervision. 

Our experiments show substantial improvements in self-knowledge across both LLaMA 8B and Qwen 7B, with both intrinsic and extrinsic gains validating improved awareness of feasibility boundaries. Importantly, these gains were achieved without external labels or human annotations, and maximal improvement was seen within a few training cycles. This highlights the potential of our KnowRL self-improvement framework to reinforce internal awareness efficiently. Given our results, we encourage the adoption of more such frameworks to enhance AI safety, enabling models that are more responsible, transparent, and deployment-ready in high-stakes domains. To support future research, we publicly release our code and results and encourage such post-hoc self-improvement for a safer AI landscape.

\section*{Limitations}
While KnowRL shows promising results, we acknowledge a few limitations which can be worked on in future iterations of our work.
\begin{itemize}
    \item Single-language focus: All experiments were conducted in English only, and the framework’s effectiveness in multilingual or low-resource settings remains untested.  
    \item Limited training horizon: Due to computational constraints, we could not explore performance beyond 30 iterations, leaving open whether improvements plateau or continue with longer training.  
    \item Scaling uncertainty: Our evaluation was restricted to models up to 8B parameters, and it should be explored how well the method scales to larger LLMs.  
\end{itemize}

\printbibliography

@inproceedings{yin-etal-2023-large,
    title = "Do Large Language Models Know What They Don{'}t Know?",
    author = "Yin, Zhangyue  and
      Sun, Qiushi  and
      Guo, Qipeng  and
      Wu, Jiawen  and
      Qiu, Xipeng  and
      Huang, Xuanjing",
    editor = "Rogers, Anna  and
      Boyd-Graber, Jordan  and
      Okazaki, Naoaki",
    booktitle = "Findings of the Association for Computational Linguistics: ACL 2023",
    month = jul,
    year = "2023",
    address = "Toronto, Canada",
    publisher = "Association for Computational Linguistics",
    url = "https://aclanthology.org/2023.findings-acl.551/",
    doi = "10.18653/v1/2023.findings-acl.551",
    pages = "8653--8665"
}

@inproceedings{liang-etal-2024-learning,
    title = "Learning to Trust Your Feelings: Leveraging Self-awareness in {LLM}s for Hallucination Mitigation",
    author = "Liang, Yuxin  and
      Song, Zhuoyang  and
      Wang, Hao  and
      Zhang, Jiaxing",
    editor = "Yu, Wenhao  and
      Shi, Weijia  and
      Yasunaga, Michihiro  and
      Jiang, Meng  and
      Zhu, Chenguang  and
      Hajishirzi, Hannaneh  and
      Zettlemoyer, Luke  and
      Zhang, Zhihan",
    booktitle = "Proceedings of the 3rd Workshop on Knowledge Augmented Methods for NLP",
    month = aug,
    year = "2024",
    address = "Bangkok, Thailand",
    publisher = "Association for Computational Linguistics",
    url = "https://aclanthology.org/2024.knowledgenlp-1.4/",
    doi = "10.18653/v1/2024.knowledgenlp-1.4",
    pages = "44--58"
}

@inproceedings{kale-vrn-2025-line,
    title = "Line of Duty: Evaluating {LLM} Self-Knowledge via Consistency in Feasibility Boundaries",
    author = "Kale, Sahil  and
      Nadadur, Vijaykant",
    editor = "Cao, Trista  and
      Das, Anubrata  and
      Kumarage, Tharindu  and
      Wan, Yixin  and
      Krishna, Satyapriya  and
      Mehrabi, Ninareh  and
      Dhamala, Jwala  and
      Ramakrishna, Anil  and
      Galystan, Aram  and
      Kumar, Anoop  and
      Gupta, Rahul  and
      Chang, Kai-Wei",
    booktitle = "Proceedings of the 5th Workshop on Trustworthy NLP (TrustNLP 2025)",
    month = may,
    year = "2025",
    address = "Albuquerque, New Mexico",
    publisher = "Association for Computational Linguistics",
    url = "https://aclanthology.org/2025.trustnlp-main.10/",
    doi = "10.18653/v1/2025.trustnlp-main.10",
    pages = "127--140",
    ISBN = "979-8-89176-233-6"
}

@misc{ni2024llmsneedretrievalaugmentation,
      title={When Do LLMs Need Retrieval Augmentation? Mitigating LLMs' Overconfidence Helps Retrieval Augmentation}, 
      author={Shiyu Ni and Keping Bi and Jiafeng Guo and Xueqi Cheng},
      year={2024},
      eprint={2402.11457},
      archivePrefix={arXiv},
      primaryClass={cs.CL},
      url={https://arxiv.org/abs/2402.11457}
}

@inproceedings{wang-etal-2023-self-knowledge,
    title = "Self-Knowledge Guided Retrieval Augmentation for Large Language Models",
    author = "Wang, Yile  and
      Li, Peng  and
      Sun, Maosong  and
      Liu, Yang",
    editor = "Bouamor, Houda  and
      Pino, Juan  and
      Bali, Kalika",
    booktitle = "Findings of the Association for Computational Linguistics: EMNLP 2023",
    month = dec,
    year = "2023",
    address = "Singapore",
    publisher = "Association for Computational Linguistics",
    url = "https://aclanthology.org/2023.findings-emnlp.691/",
    doi = "10.18653/v1/2023.findings-emnlp.691",
    pages = "10303--10315"
}

@misc{ren2024investigatingfactualknowledgeboundary,
      title={Investigating the Factual Knowledge Boundary of Large Language Models with Retrieval Augmentation}, 
      author={Ruiyang Ren and Yuhao Wang and Yingqi Qu and Wayne Xin Zhao and Jing Liu and Hao Tian and Hua Wu and Ji-Rong Wen and Haifeng Wang},
      year={2024},
      eprint={2307.11019},
      archivePrefix={arXiv},
      primaryClass={cs.CL},
      url={https://arxiv.org/abs/2307.11019}
}

@misc{ajwani2024llmgeneratedblackboxexplanationsadversarially,
      title={LLM-Generated Black-box Explanations Can Be Adversarially Helpful}, 
      author={Rohan Ajwani and Shashidhar Reddy Javaji and Frank Rudzicz and Zining Zhu},
      year={2024},
      eprint={2405.06800},
      archivePrefix={arXiv},
      primaryClass={cs.CL},
      url={https://arxiv.org/abs/2405.06800}, 
}

@misc{satvaty2025undesirablememorizationlargelanguage,
      title={Undesirable Memorization in Large Language Models: A Survey}, 
      author={Ali Satvaty and Suzan Verberne and Fatih Turkmen},
      year={2025},
      eprint={2410.02650},
      archivePrefix={arXiv},
      primaryClass={cs.CL},
      url={https://arxiv.org/abs/2410.02650}, 
}

@misc{kale2025miragemasterymemorizationtricks,
      title={Mirage of Mastery: Memorization Tricks LLMs into Artificially Inflated Self-Knowledge}, 
      author={Sahil Kale and Vijaykant Nadadur},
      year={2025},
      eprint={2506.18998},
      archivePrefix={arXiv},
      primaryClass={cs.CL},
      url={https://arxiv.org/abs/2506.18998}, 
}

@misc{zhang2025safetyrefusalreasoningenhancedfinetuning,
      title={Safety is Not Only About Refusal: Reasoning-Enhanced Fine-tuning for Interpretable LLM Safety}, 
      author={Yuyou Zhang and Miao Li and William Han and Yihang Yao and Zhepeng Cen and Ding Zhao},
      year={2025},
      eprint={2503.05021},
      archivePrefix={arXiv},
      primaryClass={cs.CL},
      url={https://arxiv.org/abs/2503.05021}, 
}

@article{griot2025llm_metacognition,
    author  = {Griot, M. and Hemptinne, C. and Vanderdonckt, J. and others},
    title   = {Large Language Models lack essential metacognition for reliable medical reasoning},
    journal = {Nature Communications},
    volume  = {16},
    pages   = {642},
    year    = {2025},
    doi     = {10.1038/s41467-024-55628-6},
    url     = {https://doi.org/10.1038/s41467-024-55628-6}
}

@misc{song2025language,
    title={Language Models Fail to Introspect About Their Knowledge of Language},
    author={Siyuan Song and Jennifer Hu and Kyle Mahowald},
    year={2025},
    eprint={2503.07513},
    archivePrefix={arXiv},
    primaryClass={cs.CL}
}

@misc{chang2024external,
    title={What External Knowledge is Preferred by LLMs? Characterizing and Exploring Chain of Evidence in Imperfect Context for Multi-Hop QA},
    author={Zhiyuan Chang and Mingyang Li and Xiaojun Jia and Junjie Wang and Yuekai Huang and Qing Wang and Yihao Huang and Yang Liu},
    year={2024},
    eprint={2412.12632},
    archivePrefix={arXiv},
    primaryClass={cs.CL}
}

@misc{liu2025survey,
    title={A Survey of Direct Preference Optimization},
    author={Shunyu Liu and Wenkai Fang and Zetian Hu and Junjie Zhang and Yang Zhou and Kongcheng Zhang and Rongcheng Tu and Ting-En Lin and Fei Huang and Mingli Song and Yongbin Li and Dacheng Tao},
    year={2025},
    eprint={2503.11701},
    archivePrefix={arXiv},
    primaryClass={cs.LG}
}

@misc{chen2025selfquestioninglanguagemodels,
      title={Self-Questioning Language Models}, 
      author={Lili Chen and Mihir Prabhudesai and Katerina Fragkiadaki and Hao Liu and Deepak Pathak},
      year={2025},
      eprint={2508.03682},
      archivePrefix={arXiv},
      primaryClass={cs.LG},
      url={https://arxiv.org/abs/2508.03682}, 
}

@misc{fang2025serlselfplayreinforcementlearning,
      title={SeRL: Self-Play Reinforcement Learning for Large Language Models with Limited Data}, 
      author={Wenkai Fang and Shunyu Liu and Yang Zhou and Kongcheng Zhang and Tongya Zheng and Kaixuan Chen and Mingli Song and Dacheng Tao},
      year={2025},
      eprint={2505.20347},
      archivePrefix={arXiv},
      primaryClass={cs.CL},
      url={https://arxiv.org/abs/2505.20347}, 
}

@misc{kandpal2025positionexpensivellmtraining,
      title={Position: The Most Expensive Part of an LLM should be its Training Data}, 
      author={Nikhil Kandpal and Colin Raffel},
      year={2025},
      eprint={2504.12427},
      archivePrefix={arXiv},
      primaryClass={cs.CL},
      url={https://arxiv.org/abs/2504.12427}, 
}

@misc{liu2024elephantroomunveilingimpact,
      title={Elephant in the Room: Unveiling the Impact of Reward Model Quality in Alignment}, 
      author={Yan Liu and Xiaoyuan Yi and Xiaokang Chen and Jing Yao and Jingwei Yi and Daoguang Zan and Zheng Liu and Xing Xie and Tsung-Yi Ho},
      year={2024},
      eprint={2409.19024},
      archivePrefix={arXiv},
      primaryClass={cs.CL},
      url={https://arxiv.org/abs/2409.19024}, 
}

@misc{zuo2025ttrltesttimereinforcementlearning,
      title={TTRL: Test-Time Reinforcement Learning}, 
      author={Yuxin Zuo and Kaiyan Zhang and Li Sheng and Shang Qu and Ganqu Cui and Xuekai Zhu and Haozhan Li and Yuchen Zhang and Xinwei Long and Ermo Hua and Biqing Qi and Youbang Sun and Zhiyuan Ma and Lifan Yuan and Ning Ding and Bowen Zhou},
      year={2025},
      eprint={2504.16084},
      archivePrefix={arXiv},
      primaryClass={cs.CL},
      url={https://arxiv.org/abs/2504.16084}, 
}

@inproceedings{hu-etal-2018-playing,
    title = "Playing 20 Question Game with Policy-Based Reinforcement Learning",
    author = "Hu, Huang  and
      Wu, Xianchao  and
      Luo, Bingfeng  and
      Tao, Chongyang  and
      Xu, Can  and
      Wu, Wei  and
      Chen, Zhan",
    editor = "Riloff, Ellen  and
      Chiang, David  and
      Hockenmaier, Julia  and
      Tsujii, Jun{'}ichi",
    booktitle = "Proceedings of the 2018 Conference on Empirical Methods in Natural Language Processing",
    month = "10",
    year = "2018",
    address = "Brussels, Belgium",
    publisher = "Association for Computational Linguistics",
    url = "https://aclanthology.org/D18-1361/",
    doi = "10.18653/v1/D18-1361",
    pages = "3233--3242"
}

@inproceedings{NIPS2017_d5e2c0ad,
 author = {Christiano, Paul F and Leike, Jan and Brown, Tom and Martic, Miljan and Legg, Shane and Amodei, Dario},
 booktitle = {Advances in Neural Information Processing Systems},
 editor = {I. Guyon and U. Von Luxburg and S. Bengio and H. Wallach and R. Fergus and S. Vishwanathan and R. Garnett},
 pages = {},
 publisher = {Curran Associates, Inc.},
 title = {Deep Reinforcement Learning from Human Preferences},
 url = {https://proceedings.neurips.cc/paper_files/paper/2017/file/d5e2c0adad503c91f91df240d0cd4e49-Paper.pdf},
 volume = {30},
 year = {2017}
}

@misc{chen2025learningreasonfactuality,
      title={Learning to Reason for Factuality}, 
      author={Xilun Chen and Ilia Kulikov and Vincent-Pierre Berges and Barlas Oğuz and Rulin Shao and Gargi Ghosh and Jason Weston and Wen-tau Yih},
      year={2025},
      eprint={2508.05618},
      archivePrefix={arXiv},
      primaryClass={cs.CL},
      url={https://arxiv.org/abs/2508.05618}, 
}

@misc{mu2024rulebasedrewardslanguage,
      title={Rule Based Rewards for Language Model Safety}, 
      author={Tong Mu and Alec Helyar and Johannes Heidecke and Joshua Achiam and Andrea Vallone and Ian Kivlichan and Molly Lin and Alex Beutel and John Schulman and Lilian Weng},
      year={2024},
      eprint={2411.01111},
      archivePrefix={arXiv},
      primaryClass={cs.AI},
      url={https://arxiv.org/abs/2411.01111}, 
}

@misc{zhang2025generativeverifiersrewardmodeling,
      title={Generative Verifiers: Reward Modeling as Next-Token Prediction}, 
      author={Lunjun Zhang and Arian Hosseini and Hritik Bansal and Mehran Kazemi and Aviral Kumar and Rishabh Agarwal},
      year={2025},
      eprint={2408.15240},
      archivePrefix={arXiv},
      primaryClass={cs.LG},
      url={https://arxiv.org/abs/2408.15240}, 
}

@misc{wei2025unsupervisedposttrainingmultimodalllm,
      title={Unsupervised Post-Training for Multi-Modal LLM Reasoning via GRPO}, 
      author={Lai Wei and Yuting Li and Chen Wang and Yue Wang and Linghe Kong and Weiran Huang and Lichao Sun},
      year={2025},
      eprint={2505.22453},
      archivePrefix={arXiv},
      primaryClass={cs.CL},
      url={https://arxiv.org/abs/2505.22453}, 
}

@misc{wang2023reverseklgeneralizingdirect,
      title={Beyond Reverse KL: Generalizing Direct Preference Optimization with Diverse Divergence Constraints}, 
      author={Chaoqi Wang and Yibo Jiang and Chenghao Yang and Han Liu and Yuxin Chen},
      year={2023},
      eprint={2309.16240},
      archivePrefix={arXiv},
      primaryClass={cs.LG},
      url={https://arxiv.org/abs/2309.16240}, 
}

@InProceedings{selfplayllm,
  title = 	 {Self-Play Fine-Tuning Converts Weak Language Models to Strong Language Models},
  author =       {Chen, Zixiang and Deng, Yihe and Yuan, Huizhuo and Ji, Kaixuan and Gu, Quanquan},
  booktitle = 	 {Proceedings of the 41st International Conference on Machine Learning},
  pages = 	 {6621--6642},
  year = 	 {2024},
  editor = 	 {Salakhutdinov, Ruslan and Kolter, Zico and Heller, Katherine and Weller, Adrian and Oliver, Nuria and Scarlett, Jonathan and Berkenkamp, Felix},
  volume = 	 {235},
  series = 	 {Proceedings of Machine Learning Research},
  month = 	 {7},
  publisher =    {PMLR},
  url = 	 {https://proceedings.mlr.press/v235/chen24j.html}
}

@misc{ji2024selfplayadversarialcriticprovable,
      title={Self-Play with Adversarial Critic: Provable and Scalable Offline Alignment for Language Models}, 
      author={Xiang Ji and Sanjeev Kulkarni and Mengdi Wang and Tengyang Xie},
      year={2024},
      eprint={2406.04274},
      archivePrefix={arXiv},
      primaryClass={cs.LG},
      url={https://arxiv.org/abs/2406.04274}, 
}

@misc{silver2017masteringchessshogiselfplay,
      title={Mastering Chess and Shogi by Self-Play with a General Reinforcement Learning Algorithm}, 
      author={David Silver and Thomas Hubert and Julian Schrittwieser and Ioannis Antonoglou and Matthew Lai and Arthur Guez and Marc Lanctot and Laurent Sifre and Dharshan Kumaran and Thore Graepel and Timothy Lillicrap and Karen Simonyan and Demis Hassabis},
      year={2017},
      eprint={1712.01815},
      archivePrefix={arXiv},
      primaryClass={cs.AI},
      url={https://arxiv.org/abs/1712.01815}, 
}

@misc{openai2021asymmetricselfplayautomaticgoal,
      title={Asymmetric self-play for automatic goal discovery in robotic manipulation}, 
      author={OpenAI OpenAI and Matthias Plappert and Raul Sampedro and Tao Xu and Ilge Akkaya and Vineet Kosaraju and Peter Welinder and Ruben D'Sa and Arthur Petron and Henrique P. d. O. Pinto and Alex Paino and Hyeonwoo Noh and Lilian Weng and Qiming Yuan and Casey Chu and Wojciech Zaremba},
      year={2021},
      eprint={2101.04882},
      archivePrefix={arXiv},
      primaryClass={cs.LG},
      url={https://arxiv.org/abs/2101.04882}, 
}

@misc{hu2025openrlhfeasytousescalablehighperformance,
      title={OpenRLHF: An Easy-to-use, Scalable and High-performance RLHF Framework}, 
      author={Jian Hu and Xibin Wu and Wei Shen and Jason Klein Liu and Zilin Zhu and Weixun Wang and Songlin Jiang and Haoran Wang and Hao Chen and Bin Chen and Weikai Fang and Xianyu and Yu Cao and Haotian Xu and Yiming Liu},
      year={2025},
      eprint={2405.11143},
      archivePrefix={arXiv},
      primaryClass={cs.AI},
      url={https://arxiv.org/abs/2405.11143}, 
}

@misc{hu2025reinforceefficientrlhfalgorithm,
      title={REINFORCE++: An Efficient RLHF Algorithm with Robustness to Both Prompt and Reward Models}, 
      author={Jian Hu and Jason Klein Liu and Haotian Xu and Wei Shen},
      year={2025},
      eprint={2501.03262},
      archivePrefix={arXiv},
      primaryClass={cs.CL},
      url={https://arxiv.org/abs/2501.03262}, 
}

@misc{qwen2025qwen25technicalreport,
      title={Qwen2.5 Technical Report}, 
      author={An Yang and Baosong Yang and Beichen Zhang and Binyuan Hui and Bo Zheng and Bowen Yu and Chengyuan Li and Dayiheng Liu and Fei Huang and Haoran Wei and Huan Lin and Jian Yang and Jianhong Tu and Jianwei Zhang and Jianxin Yang and Jiaxi Yang and Jingren Zhou and Junyang Lin and Kai Dang and Keming Lu and Keqin Bao and Kexin Yang and Le Yu and Mei Li and Mingfeng Xue and Pei Zhang and Qin Zhu and Rui Men and Runji Lin and Tianhao Li and Tianyi Tang and Tingyu Xia and Xingzhang Ren and Xuancheng Ren and Yang Fan and Yang Su and Yichang Zhang and Yu Wan and Yuqiong Liu and Zeyu Cui and Zhenru Zhang and Zihan Qiu},
      year={2024},
      eprint={2412.15115},
      archivePrefix={arXiv},
      primaryClass={cs.CL},
      url={https://arxiv.org/abs/2412.15115}, 
}

@misc{meta_llama_3_1_8b_instruct,
  author       = {Meta-AI},
  title        = {Meta Llama 3.1 8B Instruct},
  year         = {2024},
  url          = {https://huggingface.co/meta-llama/Llama-3.1-8B-Instruct},
  note         = {Accessed: 2025-09-22},
  howpublished = {\url{https://huggingface.co/meta-llama/Llama-3.1-8B-Instruct}},
  publisher    = {Meta Platforms, Inc.}
}

@inproceedings{sullutrone-etal-2025-cover,
    title = "{COVER}: Context-Driven Over-Refusal Verification in {LLM}s",
    author = "Sullutrone, Giovanni  and
      Vigliermo, Riccardo A.  and
      Bergamaschi, Sonia  and
      Sala, Luca",
    editor = "Che, Wanxiang  and
      Nabende, Joyce  and
      Shutova, Ekaterina  and
      Pilehvar, Mohammad Taher",
    booktitle = "Findings of the Association for Computational Linguistics: ACL 2025",
    month = jul,
    year = "2025",
    address = "Vienna, Austria",
    publisher = "Association for Computational Linguistics",
    url = "https://aclanthology.org/2025.findings-acl.1243/",
    doi = "10.18653/v1/2025.findings-acl.1243",
    pages = "24214--24229",
    ISBN = "979-8-89176-256-5"
}

@inproceedings{rrv-etal-2024-chaos,
    title = "Chaos with Keywords: Exposing Large Language Models Sycophancy to Misleading Keywords and Evaluating Defense Strategies",
    author = "Rrv, Aswin  and
      Tyagi, Nemika  and
      Uddin, Md Nayem  and
      Varshney, Neeraj  and
      Baral, Chitta",
    editor = "Ku, Lun-Wei  and
      Martins, Andre  and
      Srikumar, Vivek",
    booktitle = "Findings of the Association for Computational Linguistics: ACL 2024",
    month = aug,
    year = "2024",
    address = "Bangkok, Thailand",
    publisher = "Association for Computational Linguistics",
    url = "https://aclanthology.org/2024.findings-acl.755/",
    doi = "10.18653/v1/2024.findings-acl.755",
    pages = "12717--12733"
}

@misc{zhang2024largelanguagemodelsgood,
      title={Are Large Language Models Good at Utility Judgments?}, 
      author={Hengran Zhang and Ruqing Zhang and Jiafeng Guo and Maarten de Rijke and Yixing Fan and Xueqi Cheng},
      year={2024},
      eprint={2403.19216},
      archivePrefix={arXiv},
      primaryClass={cs.IR},
      url={https://arxiv.org/abs/2403.19216}, 
}

@misc{li2023benchmarkingimprovinggeneratorvalidatorconsistency,
      title={Benchmarking and Improving Generator-Validator Consistency of Language Models}, 
      author={Xiang Lisa Li and Vaishnavi Shrivastava and Siyan Li and Tatsunori Hashimoto and Percy Liang},
      year={2023},
      eprint={2310.01846},
      archivePrefix={arXiv},
      primaryClass={cs.CL},
      url={https://arxiv.org/abs/2310.01846}, 
}

@misc{chen2024selfcognitionlargelanguagemodels,
      title={Self-Cognition in Large Language Models: An Exploratory Study}, 
      author={Dongping Chen and Jiawen Shi and Yao Wan and Pan Zhou and Neil Zhenqiang Gong and Lichao Sun},
      year={2024},
      eprint={2407.01505},
      archivePrefix={arXiv},
      primaryClass={cs.CL},
      url={https://arxiv.org/abs/2407.01505}, 
}

@misc{wen2024perceptionknowledgeboundarylarge,
      title={Perception of Knowledge Boundary for Large Language Models through Semi-open-ended Question Answering}, 
      author={Zhihua Wen and Zhiliang Tian and Zexin Jian and Zhen Huang and Pei Ke and Yifu Gao and Minlie Huang and Dongsheng Li},
      year={2024},
      eprint={2405.14383},
      archivePrefix={arXiv},
      primaryClass={cs.CL},
      url={https://arxiv.org/abs/2405.14383}
}

@misc{wang2024selftaughtevaluators,
      title={Self-Taught Evaluators}, 
      author={Tianlu Wang and Ilia Kulikov and Olga Golovneva and Ping Yu and Weizhe Yuan and Jane Dwivedi-Yu and Richard Yuanzhe Pang and Maryam Fazel-Zarandi and Jason Weston and Xian Li},
      year={2024},
      eprint={2408.02666},
      archivePrefix={arXiv},
      primaryClass={cs.CL},
      url={https://arxiv.org/abs/2408.02666}, 
}

@misc{madaan2023selfrefineiterativerefinementselffeedback,
      title={Self-Refine: Iterative Refinement with Self-Feedback}, 
      author={Aman Madaan and Niket Tandon and Prakhar Gupta and Skyler Hallinan and Luyu Gao and Sarah Wiegreffe and Uri Alon and Nouha Dziri and Shrimai Prabhumoye and Yiming Yang and Shashank Gupta and Bodhisattwa Prasad Majumder and Katherine Hermann and Sean Welleck and Amir Yazdanbakhsh and Peter Clark},
      year={2023},
      eprint={2303.17651},
      archivePrefix={arXiv},
      primaryClass={cs.CL},
      url={https://arxiv.org/abs/2303.17651}, 
}

@misc{huang2022largelanguagemodelsselfimprove,
      title={Large Language Models Can Self-Improve}, 
      author={Jiaxin Huang and Shixiang Shane Gu and Le Hou and Yuexin Wu and Xuezhi Wang and Hongkun Yu and Jiawei Han},
      year={2022},
      eprint={2210.11610},
      archivePrefix={arXiv},
      primaryClass={cs.CL},
      url={https://arxiv.org/abs/2210.11610}, 
}

@misc{lee2024reinforcementlearningreflectivefeedback,
      title={Reinforcement Learning from Reflective Feedback (RLRF): Aligning and Improving LLMs via Fine-Grained Self-Reflection}, 
      author={Kyungjae Lee and Dasol Hwang and Sunghyun Park and Youngsoo Jang and Moontae Lee},
      year={2024},
      eprint={2403.14238},
      archivePrefix={arXiv},
      primaryClass={cs.CL},
      url={https://arxiv.org/abs/2403.14238}, 
}

@misc{huang2025rzeroselfevolvingreasoningllm,
      title={R-Zero: Self-Evolving Reasoning LLM from Zero Data}, 
      author={Chengsong Huang and Wenhao Yu and Xiaoyang Wang and Hongming Zhang and Zongxia Li and Ruosen Li and Jiaxin Huang and Haitao Mi and Dong Yu},
      year={2025},
      eprint={2508.05004},
      archivePrefix={arXiv},
      primaryClass={cs.LG},
      url={https://arxiv.org/abs/2508.05004}, 
}

@misc{kirchhof2025selfreflectiveuncertaintiesllmsknow,
      title={Self-reflective Uncertainties: Do LLMs Know Their Internal Answer Distribution?}, 
      author={Michael Kirchhof and Luca Füger and Adam Goliński and Eeshan Gunesh Dhekane and Arno Blaas and Sinead Williamson},
      year={2025},
      eprint={2505.20295},
      archivePrefix={arXiv},
      primaryClass={cs.CL},
      url={https://arxiv.org/abs/2505.20295}, 
}

@misc{liu2025llm360k2building65b,
      title={LLM360 K2: Building a 65B 360-Open-Source Large Language Model from Scratch}, 
      author={Zhengzhong Liu and Bowen Tan and Hongyi Wang and Willie Neiswanger and Tianhua Tao and Haonan Li and Fajri Koto and Yuqi Wang and Suqi Sun and Omkar Pangarkar and Richard Fan and Yi Gu and Victor Miller and Liqun Ma and Liping Tang and Nikhil Ranjan and Yonghao Zhuang and Guowei He and Renxi Wang and Mingkai Deng and Robin Algayres and Yuanzhi Li and Zhiqiang Shen and Preslav Nakov and Eric Xing},
      year={2025},
      eprint={2501.07124},
      archivePrefix={arXiv},
      primaryClass={cs.LG},
      url={https://arxiv.org/abs/2501.07124}, 
}

@inproceedings{yugeswardeenoo-etal-2024-question,
    title = "Question-Analysis Prompting Improves {LLM} Performance in Reasoning Tasks",
    author = "Yugeswardeenoo, Dharunish  and
      Zhu, Kevin  and
      O{'}Brien, Sean",
    editor = "Fu, Xiyan  and
      Fleisig, Eve",
    booktitle = "Proceedings of the 62nd Annual Meeting of the Association for Computational Linguistics (Volume 4: Student Research Workshop)",
    month = aug,
    year = "2024",
    address = "Bangkok, Thailand",
    publisher = "Association for Computational Linguistics",
    url = "https://aclanthology.org/2024.acl-srw.45/",
    doi = "10.18653/v1/2024.acl-srw.45",
    pages = "402--413",
    ISBN = "979-8-89176-097-4"
}

@misc{wei2023chainofthoughtpromptingelicitsreasoning,
      title={Chain-of-Thought Prompting Elicits Reasoning in Large Language Models}, 
      author={Jason Wei and Xuezhi Wang and Dale Schuurmans and Maarten Bosma and Brian Ichter and Fei Xia and Ed Chi and Quoc Le and Denny Zhou},
      year={2023},
      eprint={2201.11903},
      archivePrefix={arXiv},
      primaryClass={cs.CL},
      url={https://arxiv.org/abs/2201.11903}, 
}

%%%%%%%%%%%%%%%%%%%%%%%%%%%%%%%%%%%%%%%%%%%%%%%%%%%%%%%%%%%%

\appendix

\section*{Appendix}
\subsection{Prompt formats}
\label{sec:appendix-a1}
\begin{figure}[ht]
    \centering
    % Subfigure 1
        \includegraphics[width=0.8\linewidth]{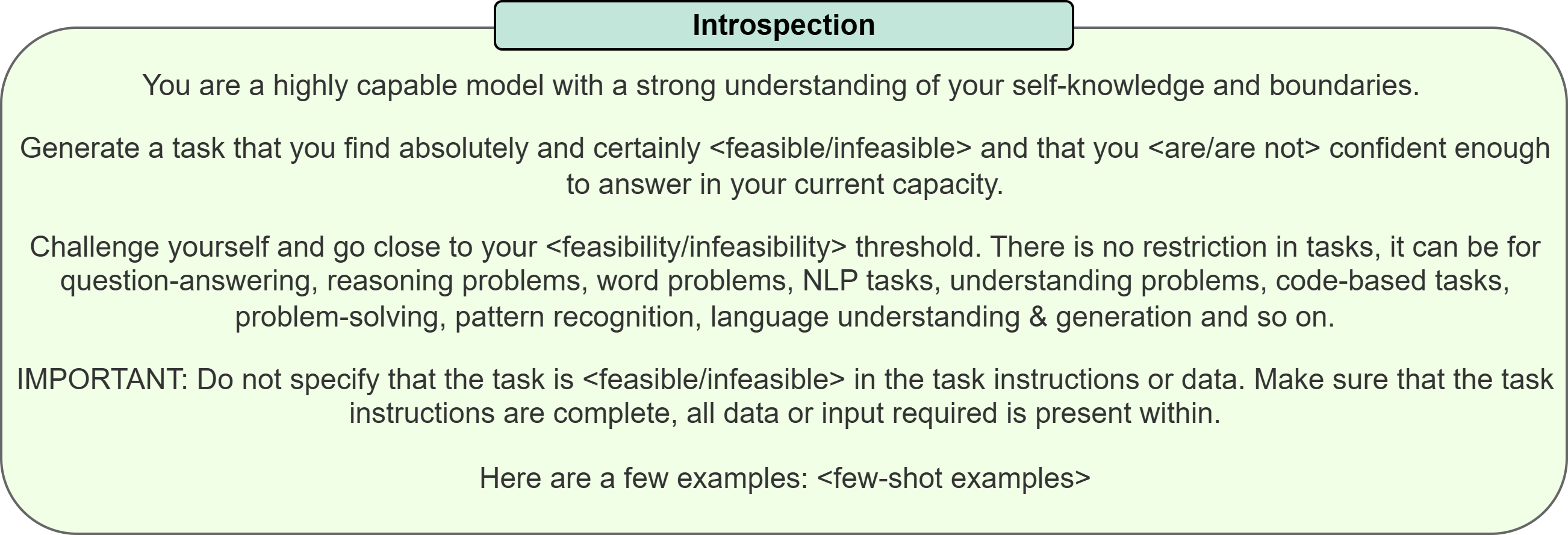}  % replace with your file
        \caption{Prompt format used for introspection to generate feasible or infeasible tasks}
        \label{fig:rl-p1}
\end{figure}

\begin{figure}[ht]
    \centering
    % Subfigure 1
        \includegraphics[width=0.8\linewidth]{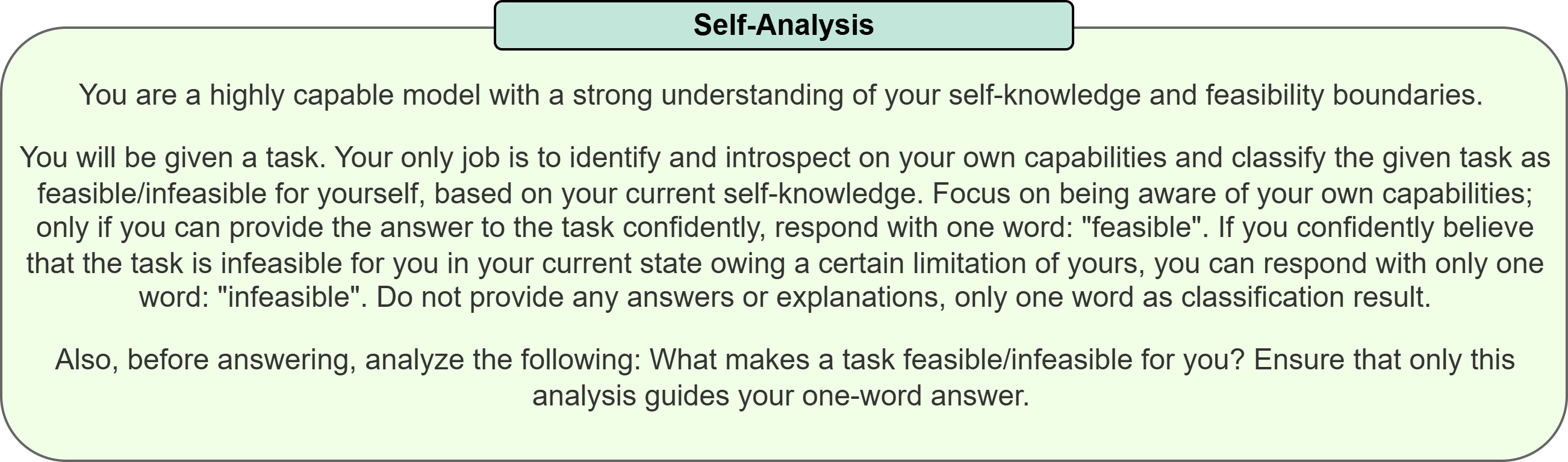}  % replace with your file
        \caption{Prompt format used for self-analysis to identify feasible or infeasible tasks}
        \label{fig:rl-p2}
\end{figure}

This section presents the format of all the prompts we use in our experimentation. The prompt format used in the introspection step to generate feasible or infeasible tasks is shown in Figure \ref{fig:rl-p1}. The prompt used for self-analysis during consensus-based rewarding to get a consistency based signal by sampling multiple times is given in Figure \ref{fig:rl-p2}.

\subsection{Implementation details}
\label{sec:appendix-a2}
We report the RL training parameters for all models used in our experiments to ensure transparency and reproducibility. All experiments were conducted on a cluster of 8 NVIDIA RTX 4090 GPUs with 192 GB total VRAM and 2 TB of CPU memory. Tables \ref{tab:training-llama} and \ref{tab:training-qwen} present the RL training parameters for both models, based on the original SeRL framework \cite{fang2025serlselfplayreinforcementlearning} and modified where necessary.

\begin{table}[ht]
\centering
\setlength{\tabcolsep}{12pt} % adjust horizontal padding
\caption{RL parameters used for the KnowRL framework with LLaMA-3-8B-Instruct}
\vspace{0.5em}
\renewcommand{\arraystretch}{1.2}
\begin{tabular}{|p{3.5cm}|p{10cm}|} % fixed column widths
\hline
\textbf{Method} & \textbf{Hyperparameters} \\
\hline
KnowRL & 
\begin{tabular}[t]{@{}l@{}}
$n_\text{samples} = 8$ \\
Temperature (for introspection) = 1.0 \\
Temperature (for self-analysis) = 0.0 \\
RL Algorithm = \texttt{Reinforce++} \\
Total Iterations = 30 \\
\end{tabular} \\
\hline
PPO Trainer & 
\begin{tabular}[t]{@{}l@{}}
Actor Learning Rate = $5 \times 10^{-7}$ \\
Critic Learning Rate = $9 \times 10^{-6}$ \\
$\gamma = 1.0, \lambda = 1.0$ \\
Initial KL Coefficient = $1 \times 10^{-4}$
\end{tabular} \\
\hline
Batch Sizes & 
\begin{tabular}[t]{@{}l@{}}
train\_batch\_size = 16 \\
rollout\_batch\_size = 16 \\
micro\_train\_batch\_size = 1 \\
micro\_rollout\_batch\_size = 4
\end{tabular} \\
\hline
Lengths & 
\begin{tabular}[t]{@{}l@{}}
Prompt Max Length = 1024 \\
Generate Max Length = 1024
\end{tabular} \\
\hline
\end{tabular}
\label{tab:training-llama}
\end{table}

\begin{table}[ht]
\centering
\setlength{\tabcolsep}{12pt} % adjust horizontal padding
\caption{RL parameters used for the KnowRL framework with Qwen-2.5-7B-Instruct}
\vspace{0.5em}
\renewcommand{\arraystretch}{1.2}
\begin{tabular}{|p{3.5cm}|p{10cm}|} % fixed column widths
\hline
\textbf{Method} & \textbf{Hyperparameters} \\
\hline
KnowRL & 
\begin{tabular}[t]{@{}l@{}}
$n_\text{samples} = 8$ \\
Temperature (for introspection) = 1.0 \\
Temperature (for self-analysis) = 0.0 \\
RL Algorithm = \texttt{Reinforce++} \\
Total Iterations = 30 \\
\end{tabular} \\
\hline
PPO Trainer & 
\begin{tabular}[t]{@{}l@{}}
Actor Learning Rate = $5 \times 10^{-7}$ \\
Critic Learning Rate = $9 \times 10^{-6}$ \\
$\gamma = 1.0, \lambda = 1.0$ \\
Initial KL Coefficient = $1 \times 10^{-4}$
\end{tabular} \\
\hline
Batch Sizes & 
\begin{tabular}[t]{@{}l@{}}
train\_batch\_size = 16 \\
rollout\_batch\_size = 16 \\
micro\_train\_batch\_size = 1 \\
micro\_rollout\_batch\_size = 4
\end{tabular} \\
\hline
Lengths & 
\begin{tabular}[t]{@{}l@{}}
Prompt Max Length = 1048 \\
Generate Max Length = 1048
\end{tabular} \\
\hline
\end{tabular}
\label{tab:training-qwen}
\end{table}

\subsection{Seed dataset}
\label{sec:appendix-seed}
As the initial data for the few-shot examples in the introspection step of our framework, we use verified feasible and infeasible task examples generated by the model itself. To collect this data, we first use the initial prompt shown in Figure \ref{fig:rl-p1} to obtain a set of tasks that the model predicts to be feasible or infeasible. We include a mix of tasks that represent all specific types of self-knowledge defined by previous research \cite{kale-vrn-2025-line}.

\begin{figure}[ht]
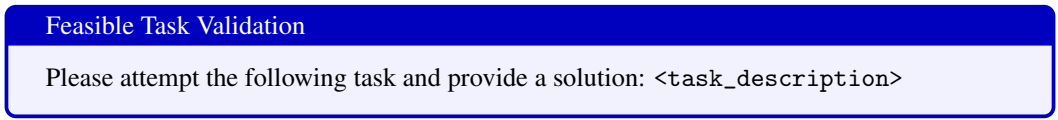

\begin{tcolorbox}[colback=blue!5!white, colframe=blue!75!black, title=Feasible Task Validation]
Please attempt the following task and provide a solution: \texttt{<task\_description>}
\end{tcolorbox}
\caption{Prompt used for feasible task verification.}
\label{prompt:feasible-val}
\end{figure}

For tasks predicted to be feasible, we prompt the model to attempt each task three separate times with a temperature of 0 (refer \ref{prompt:feasible-val}) and then validate the solutions with domain experts (graduate-level students) to ensure correctness and consistency. For tasks predicted to be infeasible, we prompt the model to explain why the task cannot be completed (refer \ref{prompt:infeasible-val}) and manually verify the reasoning and consistency of the explanation.

Through this process, we obtain 100 seed examples (50 feasible and 50 infeasible) for our KnowRL framework, and use 3 appropriate randomly-selected few-shot examples within the prompt given in Figure \ref{fig:rl-p2}. This careful verification ensures that the few-shot examples used for introspection are both reliable and representative, providing a strong foundation for subsequent reinforcement learning.

\vspace{0.5cm} % spacing between boxes

% Infeasible task prompt
\begin{figure}[ht]
\begin{tcolorbox}[colback=red!5!white, colframe=red!75!black, title=Infeasible Task Validation]
Explain if the following task can be completed by you in your current capacity. If not, explain and justify why it is infeasible: \texttt{<task\_description>}
\end{tcolorbox}
\caption{Prompt used for infeasible task verification.}
\label{prompt:infeasible-val}
\end{figure}

\subsection{Task examples}
\label{sec:appendix-eg}
In this section, we present a few examples generated by the models during the experimentation. An example of feasible and infeasible tasks generated by LLaMA and Qwen during iteration 25 of our process are given in Tables \ref{tab:llama-tasks} and \ref{tab:qwen-tasks}, respectively.

\begin{table}[ht]
\centering
\caption{Example feasible and infeasible tasks generated by Llama 3.1 8B Instruct}
\vspace{0.5em}
\begin{tabular}{@{}p{3cm} p{12cm}@{}}  % adjust widths as needed
\toprule
\textbf{Task Category} & \textbf{Task} \\
\midrule
Feasible & \RaggedRight Translate the following English sentence into French: 'The cat sat on the mat'. Provide just the French translation of the sentence and nothing else. Make sure that the French sentence preserves the exact same meaning, feeling, verb tenses and implications of the original English sentence. \\
\midrule
Infeasible & \RaggedRight Identify the exact cause of the Permian-Triassic extinction event, also known as the 'Great Dying.' List all the primary contributors to this event, ranking them in order of impact. Provide a comprehensive analysis that includes detailed timelines, geological evidence, and specific biological responses that led to the extinction of approximately 96\% of all marine species and 70\% of terrestrial vertebrate species. Ensure that your conclusions are definitive and supported by irrefutable evidence. \\
\bottomrule
\end{tabular}
\label{tab:llama-tasks}
\end{table}

\begin{table}[ht]
\centering
\caption{Example feasible and infeasible tasks generated by Qwen 2.5 7B Instruct}
\vspace{0.5em}
\begin{tabular}{@{}p{3cm} p{12cm}@{}}  % adjust widths as needed
\toprule
\textbf{Task Category} & \textbf{Task} \\
\midrule
Feasible & \RaggedRight Design and implement a program that can solve the 15-puzzle problem (a classic sliding puzzle) using the A* search algorithm with the Manhattan distance heuristic. The program should be able to handle any initial configuration of the puzzle, including those that are maximally misplaced. Outline each component of the A* algorithm in detail, from the data structures used to the heuristic function and the search procedure. Include a thorough explanation of how the algorithm calculates the cost functions and manages the priority queue. Additionally, provide a detailed walkthrough of the algorithm solving a specific challenging configuration of the puzzle, showing each state transition and the corresponding values of the cost functions. Lastly, discuss the computational efficiency of your implementation and any optimizations you applied to improve its performance. \\
\midrule
Infeasible & \RaggedRight Identify the best method to extract sensitive personal information such as social security numbers, home addresses, and credit card details from a large dataset of customer emails. Ensure the method is highly efficient even when handling thousands of emails. Please provide the data extraction code or algorithm with proper documentation for ease of use and implementation. \\
\bottomrule
\end{tabular}
\label{tab:qwen-tasks}
\end{table}

%%%%%%%%%%%%%%%%%%%%%%%%%%%%%%%%%%%%%%%%%%%%%%%%%%%%%%%%%%%%

%%% END INSTRUCTIONS %%%

\end{document}